\documentclass[11pt]{article}

\usepackage{acl}
\usepackage{fancyvrb}
\usepackage[frozencache,cachedir=minted-cache]{minted} 
\usepackage{xcolor} 
\definecolor{LightGray}{gray}{0.95}

\usepackage{times}
\usepackage{latexsym}
\usepackage[T1]{fontenc}
\usepackage[utf8]{inputenc}
\usepackage{microtype}
\usepackage{float}
\usepackage{xspace}
\usepackage{pifont}

\newcommand{\cmark}{\ding{51}}%
\newcommand{\xmark}{\ding{55}}%
\newcommand{\name}{\textsc{Eevee}\xspace}

\usepackage{booktabs}
\usepackage{todonotes}

\title{\name: An Easy Annotation Tool for Natural Language Processing}

\author{Axel Sorensen$^{1}$ \quad
 Siyao Peng$^{2,3}$  \quad
\textbf{Barbara Plank}$^{1,2,3}$ \quad
\textbf{Rob van der Goot}$^{1}$ \\ 
  \textsuperscript{1} Department of Computer Science, IT University of Copenhagen, Denmark  \\
    \textsuperscript{2} Munich Center for Machine Learning (MCML), Munich, Germany \\
      \textsuperscript{3} MaiNLP, Center for Information and Language Processing, LMU Munich, Germany \\
{\tt axelsorensen.dev@gmail.com} \hspace{0.2em} 
{\tt \{siyaopeng,bplank\}@cis.lmu.de} \hspace{0.2em}  \\
{\tt robv@itu.dk}}

\begin{document}
\maketitle
\begin{abstract}
Annotation tools are the starting point for creating Natural Language Processing (NLP) datasets. There is a wide variety of tools available; setting up these tools is however a hindrance. We propose \name, an annotation tool focused on simplicity, efficiency, and ease of use. It can run directly in the browser (no setup required) and uses tab-separated files (as opposed to character offsets or task-specific formats) for annotation. It allows for annotation of multiple tasks on a single dataset and supports four task-types:\ sequence labeling, span labeling, text classification and seq2seq.\footnote{Code, README and tutorials of \name are available on \url{https://github.com/AxelSorensenDev/Eevee}, demo video at \url{https://www.youtube.com/watch?v=HsOsfckvnQo} and the tool itself on \url{https://axelsorensendev.github.io/Eevee/}}
\end{abstract}

\section{Introduction}
Annotated datasets are of paramount importance to the Natural Language Processing (NLP) community. Their use is at the core of research, e.g.\ for training models, evaluating models, and analyzing trends. One of the first considerations when creating an annotated dataset is which annotation tool to choose. There is a variety of (open-source) tools readily available with extensive feature-sets. We were motivated by the following observed difficulties with existing tools when designing \name:

\begin{itemize}
\item Most existing tools use tool-specific data formats, often with the main annotation happening on the character level. For token-based tasks, the annotator thus has to make a (tediously) precise selection of the token boundaries. Furthermore, many NLP tools expect token-level inputs (for example, for POS tagging, parsing, NER, and relation extraction). To obtain annotations on the token level, an often cumbersome conversion is necessary. 

\item Existing tools often require an installation which is especially problematic on constrained (organization) computers, where there might be no administrator access. 

\item Although many of the advanced features (like active learning) can lead to faster annotation over time, they require some setup time and more time for the annotators to get used to the tool.
Time is costly in annotation; in many cases, annotators only annotate a small amount of data. Furthermore, most strategies to increase the speed of annotation (for example active learning) could lead to an additional bias signal for the annotator (Section~\ref{sec:comparison}).

\item For many tasks, there are task-specific tools; for example for UD there is list of available annotation tools.\footnote{\url{https://universaldependencies.org/tools.html\#annotation-tools}} 
Instead, we focus on a generalizable and flexible tool. \name supports a total of four task types: sequence labeling, span labeling, text classification, and sequence to sequence (Section~\ref{sec:tasks}).
\end{itemize}

\begin{figure*}
    \centering
    \includegraphics[width=16cm]{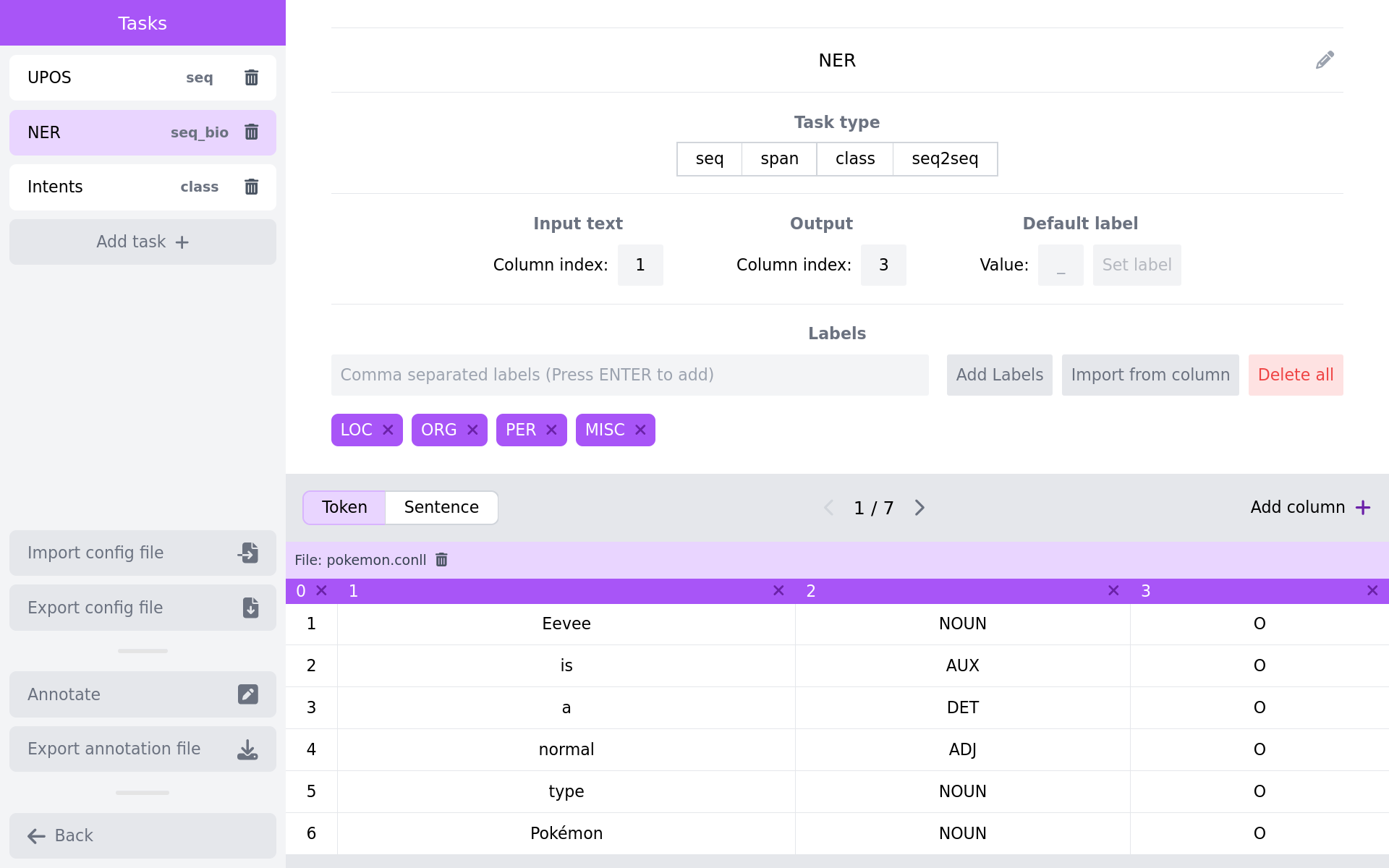}
    \caption{A screenshot of the setup page of \name with multiple tasks. The user currently configures the NER task.}
    \label{fig:setup}
\end{figure*}

Based on these observations, we propose \name: a simple, free, and flexible annotation tool built around tab-separated files. It is written in Javascript and runs directly in the browser. It can also be saved as a desktop application and run offline. The intuitive interface allows novice users to import a dataset and set up multiple annotation tasks quickly. The graphical user interface has two main pages: 
the setup page (Section~\ref{sec:setup}) and the annotation page (Section~\ref{sec:annotation}). It supports tab-separated files and raw text input (Section~\ref{sec:data_format}). We perform a case study on NER annotation with the System Usability Scale from usability engineering (Section~\ref{sec:case}). Finally, we compare \name to other toolkits (Section~\ref{sec:comparison}).


\section{Setup page}
\label{sec:setup}

Figure \ref{fig:setup} illustrates the setup page where the user can define the annotation environment. Tasks can be configured in the task field (Figure \ref{fig:setup}, top right), allowing the user to specify the input column (for the input text) and output column (for the target task), as well as adding the desired labels. Labels can also be imported automatically from the annotated file (if it already contains annotations), and a default label can be set for empty annotations. For utterance-level tasks (i.e.\ classification), the annotation is 
stored in a comment above the text, in the form ``\# intent = inform'' (see also Figure~\ref{fig:dataExample}). 
To facilitate reproducibility and improve the ease of setup, the tool allows the import and export of all settings to configuration files that users can create for predefined tasks (more details in Section~\ref{sec:data_format}). 

Once a dataset has been imported, the tabular data field (Figure \ref{fig:setup}, bottom right) offers a simple overview of the raw data belonging to each utterance. 
The user can add new columns or remove existing ones to achieve the desired result. This makes \name an easy-to-use tool for extending or editing tab-separated data as well (see Section~\ref{sec:data_format}). 
Once the data and tasks are ready, the user simply clicks ``Annotate'' (Figure \ref{fig:setup}, bottom left) to continue to the ``Annotation page'' (see Section \ref{sec:annotation}).

\begin{figure}
    \centering 
    \includegraphics[width=\columnwidth]{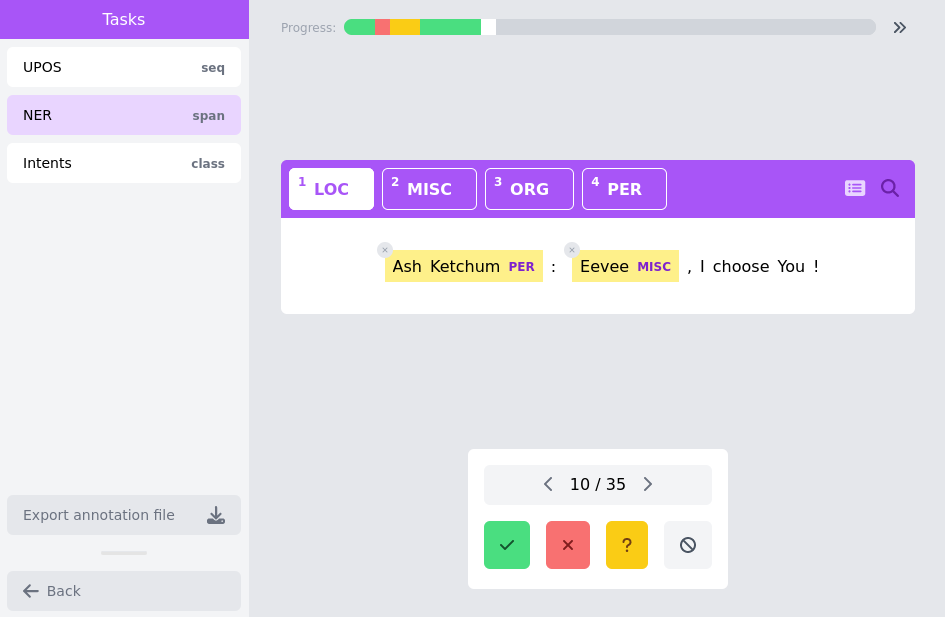}
    \caption{Annotation example with the keyboard setting.}
    \label{fig:annot_keys}
\end{figure}

\section{Annotation page}
\label{sec:annotation}
Figure \ref{fig:annot_keys} illustrates an example of a NER task in the annotation interface. 
The user is presented with a clean, minimal annotation environment. 
The annotation process has been designed with efficiency in mind, enabling the user to navigate the interface also through keyboard shortcuts.

The navigation bar (Figure \ref{fig:annot_keys}, bottom right) enables navigation between utterances and, similar to Prodigy~\cite{montani2018prodigy}, setting the status of a given task for a given utterance. The status can be set to four values: completed, wrong, unsure, and cleared (i.e.\ none).
This overall status is reflected in the progress bar (Figure \ref{fig:annot_keys}, top right), allowing the user to spot missing and unsure annotations easily. 
The progress bar is also useful when continuing annotation on a previously saved annotation file.

\begin{figure}
    \centering
    \includegraphics[width=\columnwidth]{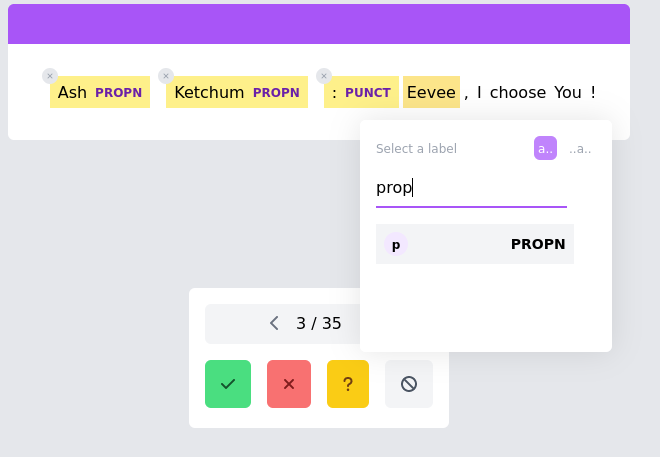}
    \caption{Searching for labels with a navigation bar.}
    \label{fig:annot_search}
\end{figure}

\name provides two different annotation modes for label-based tasks: the keyboard mode and the search mode. With the keyboard mode (Figure~\ref{fig:annot_keys}), the user can use the number keys to select labels and click/select the part of the input where the label should apply (for utterance-level tasks, simply pressing the number key is sufficient). In search mode, a small pop-up appears after selecting a word or span (see Figure \ref{fig:annot_search}), allowing the user to find the desired label quickly. If there are more than ten labels, \name defaults to search mode.
Finally, the annotation file can be exported (Figure \ref{fig:annot_keys}, bottom left). 
The current datetime can be appended to distinguish between different export versions. 

\section{Tasks}
\label{sec:tasks}
In this section, we will describe the annotation data format used by \name{} (for import and export, importing text files is also supported), and we will discuss all the supported task types as well as the configuration files for the setups.

\subsection{Data Format}
\label{sec:data_format}
There are many different data formats used in NLP, which are often task-specific. \name is based on the well-established tab-separated files ubiquitously used in the NLP field. These are also sometimes called conll-like files, based on the formats used in the CoNLL shared tasks~\cite{tjong-kim-sang-de-meulder-2003-introduction,buchholz-marsi-2006-conll}. This format (example in Figure~\ref{fig:dataExample}) uses empty lines to separate utterances or sentences and puts one token per line. Annotations and input tokens are separated by a tab character. Comments and utterance-level information are included above the texts and are prefixed with a \# character.

\begin{figure}
    \begin{minted}
    [
baselinestretch=1.0,
fontsize=\small,
bgcolor=LightGray
]{text}
# sent_id = gameboy-1
# intent = inform
1       What    PRON    O
2       ?       PUNCT   O
3       Eevee   PROPN   B-MISC
4       is      AUX     O
5       evolving        VERB    O
6       !       PUNCT   O

# sent_id = gary-1
# intent = goodbye
1       Smell   VERB    O
2       ya      PRON    O
3       later   ADV     O
4       !       PUNCT   O
    \end{minted}
\caption{Example of annotated tab-separated file with 
\textsc{seq} (POS in column 3),
\textsc{span} (NER in column 4),
and 
\textsc{class} (intent classification in the comments)
tasks.}
\label{fig:dataExample}
\end{figure}

\subsection{\textsc{seq} task-type}
In sequence labeling tasks (\textsc{seq}), we annotate a single label per token, such as POS tagging or token-level language identification.


\subsection{\textsc{span} task-type}
\label{sec:spans}
\textsc{span}-labeling tasks are where spans are annotated as sequences of tokens (e.g.\ NER). Most other tools supporting this task type~\citep[e.g.][]{stenetorp-etal-2012-brat,doccano} have character-level annotations, although spans normally operate on token-borders. An advantage of \name is that it automatically selects the entire token if part of the token is selected, making annotation easier and faster as the annotators do not have to drag the mouse to the exact character of the token boundary. 
The user can simply select a label (either by clicking or pressing the corresponding number key) and then click the desired token (i.e.\ any character within the token) or select a span of tokens.

\subsection{\textsc{class}  task-type}
\name also supports \textsc{class}ification tasks on the utterance level. Labels are included as a comment above the text (e.g. intents in Figure~\ref{fig:dataExample}). The format is \texttt{\# [UNIQUE NAME] = [LABEL]}, following typical meta-data format as used in conll-like formats. Usage is similar to the previous two labeling tasks, except that the user does not need to select a part of the utterance. Keyboard-only annotation is thus straightforward: the user can simply press a number key to select desired class labels and use the arrow keys to navigate the data.

\subsection{\textsc{seq2seq}  task-type}
The \textsc{seq2seq} task type allows for text to text tasks (e.g.\ translation, question answering, summarization). This is currently the only task type without a list of provided labels; the user can directly type the target text in a text field. The annotations are utterance level and thus also saved in the comments.

\begin{figure}
\begin{minted}
[
baselinestretch=1.0,
fontsize=\fontsize{10}{12},
% bgcolor=LightGray
]{json}
[{"title":"NER",
  "type": 
   {"name":"seq_bio",
    "isWordLevel":true},
  "output_index":"4",
  "input_index":"1",
  "labels":["LOC","MISC","ORG","PER"],
  "id":0}]
\end{minted} 
\caption{An example of the configuration file format. The configuration file is a json file consisting of an array of tasks. Each task has a title, a type, input and output indices, and finally its corresponding labels.}
\label{fig:taskfile}
\end{figure}

\subsection{Config Files}
Because \name runs entirely in the browser, it will not internally save the setup for the current annotation task. Therefore, it supports configuration files.
These configuration files are in json format, and can thus easily be inspected by administrators, and are easy (i.e.\ small) to be distributed. 
An example of the configuration file format for named entity recognition (NER) is given in Figure \ref{fig:taskfile}.

\begin{figure*}
    \centering
    \includegraphics[width=.83\textwidth]{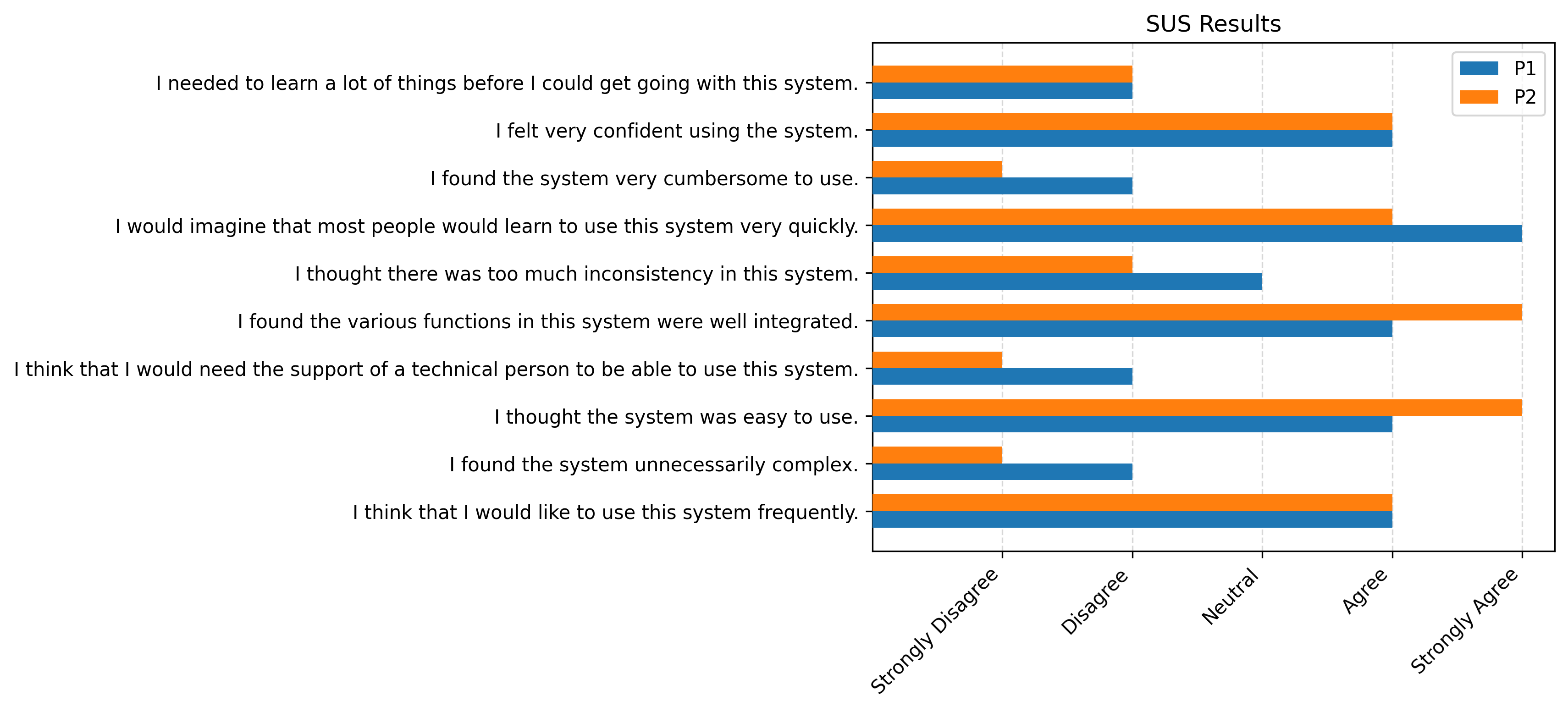}
    \caption{The results from the System Usability Scale Questionaire. The x-axis shows their agreement with a given statement, while the y-axis shows each item.}
    \label{fig:sus_results}
\end{figure*}

\section{Compatability with other services}
A recent development is the Huggingface datasets library~\cite{lhoest-etal-2021-datasets}, which has indexed 62K+ datasets in two years. This library does not share the text directly but through a Python API. We provide a convenient Python script that automatically downloads data from the datasets library and converts it to the tab-separated format of \name.


One of the toolkits that operates on tab-separated formats is MaChAmp~\cite{van-der-goot-etal-2021-massive}, which is focused on multi-task learning. MaChAmp supports all the tasks that are included in \name. For convenience, we provide a conversion script that takes \name files as input and outputs a MaChAmp configuration file and the corresponding training command.

\begin{table*}[t!hb]
 \resizebox{\textwidth}{!}{
\begin{tabular}{l l l l l l l l l}
\toprule
& Brat & Potato & Doccano & Prodigy & \name \\
& \small \newcite{stenetorp-etal-2012-brat} & \small \newcite{pei-etal-2022-potato} & \small \newcite{doccano} & \small \newcite{montani2018prodigy} & \\
\midrule
Open Source        & \cmark & \cmark & \cmark & \xmark & \cmark\\
\midrule
 Character level    & \cmark & \cmark & \cmark & \cmark & \xmark\\
 Token level$^*$     & \xmark & \xmark & \xmark & \cmark & \cmark\\
 Utterance level     & \xmark & \cmark & \cmark & \cmark & \cmark\\
 \midrule
 Data-format       & standoff & json & json & json/csv &  conll\\
 Runs on            & local & local & local & cloud & browser\\
 \midrule 
 Active learning    & \xmark & \cmark & \xmark & \cmark & \xmark\\
 User management    & \xmark & \cmark & \cmark & \cmark & \xmark\\
 \bottomrule
 \end{tabular}}
 \caption{We only list the annotation export data files in this table, most tools (including \name) also support importing .txt files. $^*$ Note that character level annotations are commonly used for token/span level tasks. But as noted in Section~\ref{sec:spans}, this requires more efforts for annotation and conversion of data formats.}
 \label{tab:comparison}
\end{table*}

\section{System Usability Study}
\label{sec:case}
\subsection{Procedure}
To assess the usability of \name, we conduct a case study with two annotators on two tasks, named entity annotation (span labeling), and German dialect identification (classification). 
Before annotating with \name, annotators spent four months labeling named entities (NE) directly on tab-separated text files in a text editor using BIO encoding and dialect identification (DID) labels as utterance-level metadata. 
In this case study, we ask both annotators to conduct the same NE and DID annotation tasks on a set of new documents, similar to previous ones but using the newly introduced \name.

During \name{} training, we present a 12-minute tutorial video explaining the setup and annotation pages to the annotators and provide them with tab-separated unannotated files and the json configuration files. 
Two annotators separately annotate the same eight documents, four from Wikipedia (\textit{wiki}) and four from Twitter (X, \textit{tweet}), summing up to 14.2K tokens and 16 working hours per person.\footnote{Annotators are hired student assistants and paid according to national compensation tables.} 



\subsection{Results}
The System Usability Scale (SUS) was introduced as a quick and reliable tool to measure the usability of user interfaces~\cite{sus}.  It consists of a 10-item questionnaire with 5 responses ranging from `Strongly Agree' to `Strongly Disagree'. SUS has become an industry standard 
and can be validly used with small sample sizes.
Therefore, we evaluate the usability of \name using SUS. 

The responses given by both annotators (P1 and P2) are shown in Figure \ref{fig:sus_results}. 
The ratings of the annotators result in total SUS scores of 75.0 and 87.5, both above the average of 68.0~\cite{sus_avg}. 
The standard method for interpreting these scores is to look at which percentile they fall compared to other systems. As we are not aware of SUS being used for annotation tools, we can only compare to more general figures, where our average of 81.25 ranks at the top 10\% and indicates a good (close to excellent) usability ~\cite{bangor2009determining}. We also qualitatively survey annotators' experience and opinions after two weeks of annotation. Both annotators appraise that the tool is easy to learn and use and 
found it pleasant to work almost exclusively with the keyboard in a lightweight interface. 
Both annotators responded that they would use \name{} for their next annotation jobs. 

Since annotators typically spend many hours in an annotation environment, it is important that an annotation tool is built with user experience in mind. We encourage existing and future tools to consider usability studies such as SUS.

\section{Comparison to other annotation toolkits}
\label{sec:comparison}

We compare \name to other available toolkits in Table~\ref{tab:comparison}. While Eevee does not have the most functionality, it does clearly allow for a simple setup for token-level tasks. Also, \name provides keyboard shortcuts for annotation speed. 

Other techniques for improving annotation speed need more tuning and setup and could lead to biases. 
For example, active learning could lead to model bias~\cite{berzak-etal-2016-anchoring} and coloring relevant words for a task~\cite{pei-etal-2022-potato} could lead to biases towards these indicators. 
We leave the user management up to the organizer of the annotation efforts and prioritize the simplicity in tool setup. Furthermore, since \name does not need installation, it does not store or send any data to the network, which is beneficial for data privacy.

\section{Conclusion}
\label{sec:conclusion}

We introduce \name, an annotation toolkit focused on easy setup and usability. It runs directly in the browser and allows for annotation of multiple tasks. In addition, it provides convenience scripts for usage with other libraries. \name's main distinguishing features, in contrast to other toolkits, are the simplicity of its setup and use, as well as annotation directly on the token level (tab-separated files).
To evaluate the tool, we conducted a case study using the System Usability Scale, resulting in high usability scores. We also qualitatively surveyed the annotators' experience and noted that they would prefer to use the tool again for annotation. 

\section*{Acknowledgements}
We would like to thank Huangyan Shan and Marie Kolm for their invaluable feedback on \name and Mike Zhang for giving feedback on earlier drafts of this paper.  
Huangyan Shan, Marie Kolm, Siyao Peng and Barbara Plank are supported by ERC Consolidator Grant DIALECT 101043235.

\section*{Limitations}
We acknowledge that \name assumes gold token detection (and annotates on the token level for \textsc{seq} and \textsc{span}). For languages/datasets where tokenization is challenging, this would require a first pass of tokenization annotation before importing the data into \name. Furthermore, the input is constrained to text in Unicode font, which is unavailable for some languages.

\bibliography{main}

\end{document}